\def\year{2021}\relax
\documentclass[letterpaper]{article} 
\usepackage{aaai21}  
\usepackage{times}  
\usepackage{helvet} 
\usepackage{courier}  
\usepackage[hyphens]{url}  
\usepackage{graphicx} 
\usepackage{subcaption}

\usepackage{color}

\usepackage{natbib}  
\usepackage{caption} 
\usepackage{multirow}
\usepackage[noend]{algpseudocode}
\usepackage{algorithmicx,algorithm}

\usepackage{amssymb}
\usepackage{pifont}

\newcommand{\cmark}{\ding{51}}%
\title{Camera-aware Proxies for Unsupervised Person Re-Identification}
\author {
    Menglin Wang\textsuperscript{\rm 1}, Baisheng Lai\textsuperscript{\rm 2}, Jianqiang Huang\textsuperscript{\rm 2}, 
    Xiaojin Gong\textsuperscript{\rm 1}\thanks{The corresponding author. This work was supported by Major Scientific Research Project of Zhejiang Lab (No. 2019DB0ZX01).}, Xian-Sheng Hua\textsuperscript{\rm 2}
    \\
}
\affiliations{
    \textsuperscript{\rm 1}Zhejiang University, Hangzhou, China  \   \  \  
    \textsuperscript{\rm 2}Alibaba Group \\
    lynnwang6875@gmail.com, baisheng.lbs@alibaba-inc.com, jianqiang.hjq@alibaba-inc.com, \\
    gongxj@zju.edu.cn, huaxiansheng@gmail.com
}

\begin{document}


\maketitle

\begin{abstract}
This paper tackles the purely unsupervised person re-identification (Re-ID) problem that requires no annotations. Some previous methods adopt clustering techniques to generate pseudo labels and use the produced labels to train Re-ID models progressively. These methods are relatively simple but effective. However, most clustering-based methods take each cluster as a pseudo identity class, neglecting the large intra-ID variance caused mainly by the change of camera views. To address this issue, we propose to split each single cluster into multiple proxies and each proxy represents the instances coming from the same camera. These camera-aware proxies enable us to deal with large intra-ID variance and generate more reliable pseudo labels for learning. Based on the camera-aware proxies, we design both intra- and inter-camera contrastive learning components for our Re-ID model to effectively learn the ID discrimination ability within and across cameras. Meanwhile, a proxy-balanced sampling strategy is also designed, which facilitates our learning further. Extensive experiments on three large-scale Re-ID datasets show that our proposed approach outperforms most unsupervised methods by a significant margin. Especially, on the challenging MSMT17 dataset, we gain $14.3\%$ Rank-1 and $10.2\%$ mAP improvements when compared to the second place. 
Code is available at: \texttt{https://github.com/Terminator8758/CAP-master}. 
\end{abstract}

\section{Introduction}
Person re-identification (Re-ID) is the task of identifying the same person in non-overlapping cameras. This task has attracted extensive research interest due to its significance in surveillance and public security. State-of-the-art Re-ID performance is achieved mainly by fully supervised methods~\cite{sun2018beyond, chen2019abd}. These methods need sufficient annotations that are expensive and time-consuming to attain, making them impractical in real-world deployments. Therefore, more and more recent studies focus on unsupervised settings, aiming to learn Re-ID models via unsupervised domain adaptation (UDA)~\cite{Wei2018PTGAN,qi2019DA,zhong2019invariance} or purely unsupervised~\cite{lin2019aBottom,li2018unsupervised,wu2019graph} techniques. Although considerable progress has been made in the unsupervised Re-ID task, there is still a large gap in performance compared to the supervised counterpart.

\begin{figure}[t]
\centering
\begin{subfigure}{0.36\textwidth}
\centering
\includegraphics[width=1.0\textwidth]{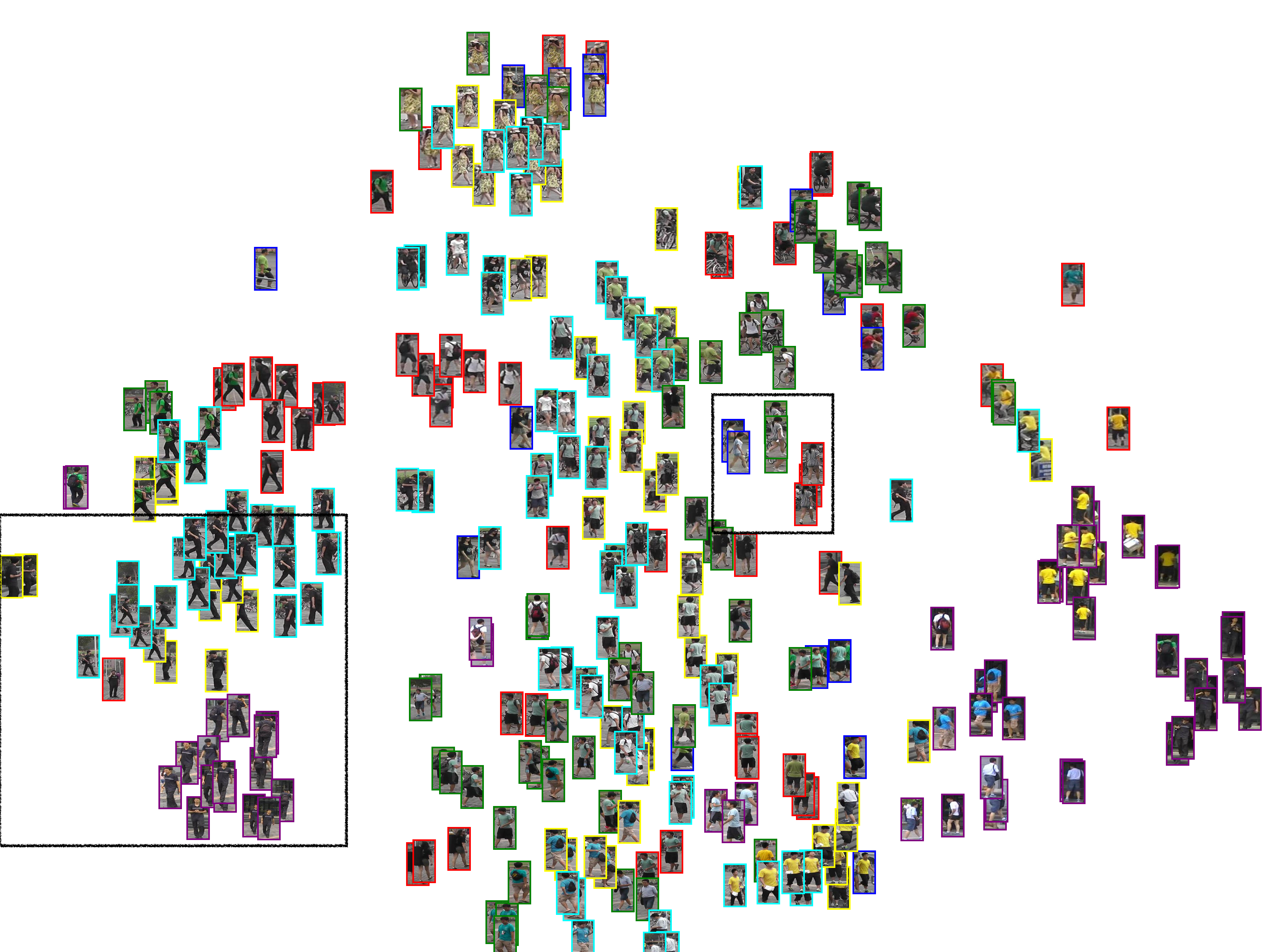} 
\caption{}
\end{subfigure}
\\
\begin{subfigure}{0.144\textwidth}
\centering
\frame{\includegraphics[width=1.0\textwidth]{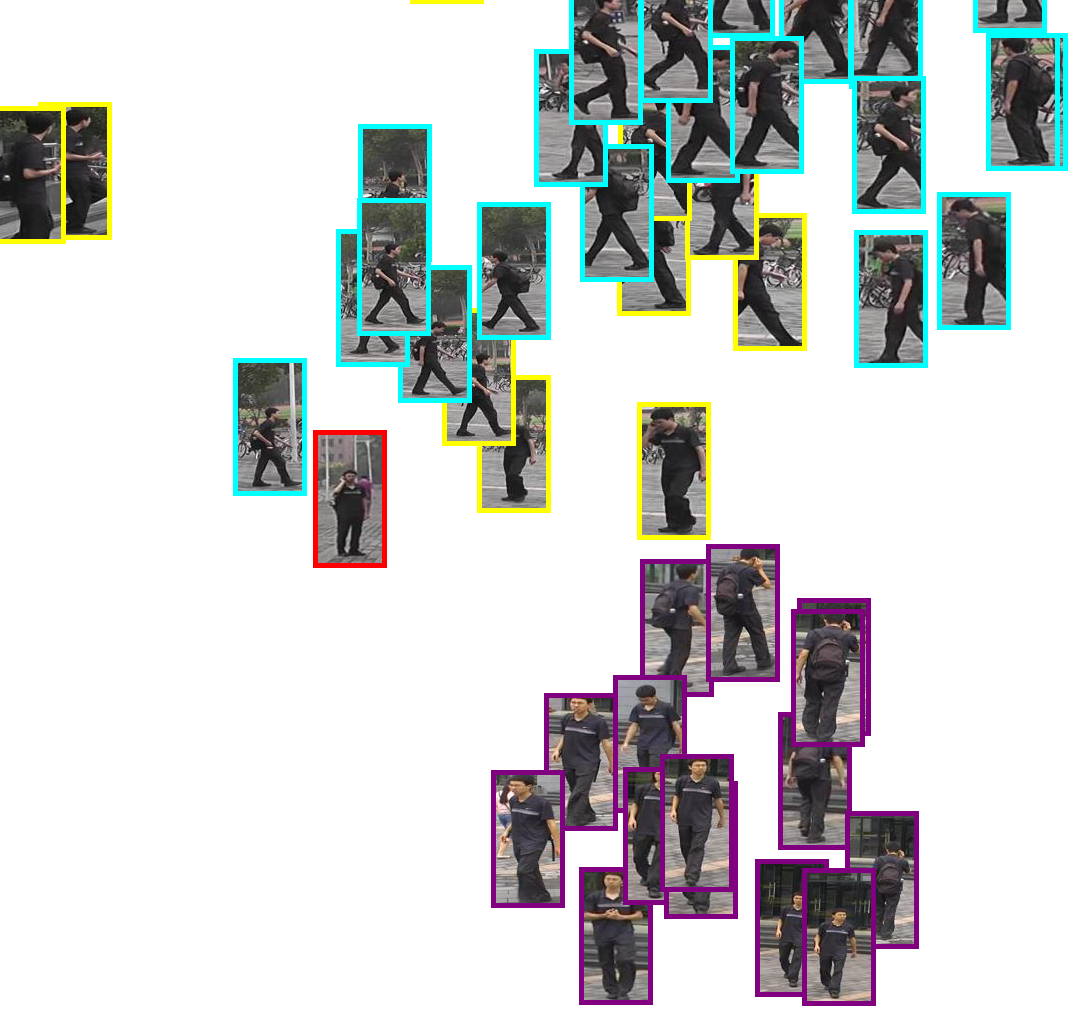}}
\caption{}
\end{subfigure}
\quad
\begin{subfigure}{0.117\textwidth}
\centering
\frame{\includegraphics[width=1.0\textwidth]{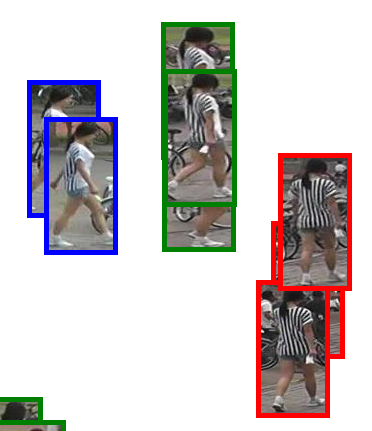}}
\caption{}
\end{subfigure}
\caption{(a) T-SNE~\cite{vanDerMaaten2008} visualization of the feature distribution on Market-1501. The features are extracted by an ImageNet-pretrained model for images of $20$ randomly selected IDs. The images from one camera are marked with the same colored bounding boxes. (b) and (c) display two sub-regions.}
\label{fig_intro}
\end{figure}

This work addresses the purely unsupervised Re-ID task, which does not require any labeled data and therefore is more challenging than the UDA-based problem. Previous methods mainly resort to pseudo labels for learning, adopting Clustering~\cite{lin2019aBottom,zeng2020hierarchical}, k-nearest neighbors (k-NN)~\cite{li2018unsupervised,chen2018deepa}, or graph~\cite{ye2017dynamic,wu2019graph} based association techniques to generate pseudo labels. The clustering-based methods learn Re-ID models by iteratively conducting a clustering step and a model updating step. These methods have a relatively simple routine but achieve promising results. Therefore, we follow this research line and propose a more effective approach.

Previous clustering-based methods~\cite{lin2019aBottom,zeng2020hierarchical,unsup_clustering,zhai2020ad} treat each cluster as a pseudo identity class, neglecting the intra-ID variance caused by the change of pose, illumination, and camera views. When observing the distribution of features extracted by an ImageNet~\cite{krizhevsky2012imagenet}-pretrained model from Market-1501~\cite{7410490}, we notice that, among the images belonging to a same ID, those within cameras are prone to gather closer than the ones from different cameras. That is, one ID may present multiple sub-clusters, as demonstrated in Figure~\ref{fig_intro}(b) and (c).

The above-mentioned phenomenon inspires us to propose a camera-aware proxy assisted learning method. Specifically, we split each single cluster, which is obtained by a camera-agnostic clustering method, into multiple camera-aware proxies. Each proxy represents the instances coming from the same camera. These camera-aware proxies can better capture local structures within IDs. More important, when treating each proxy as an intra-camera pseudo identity class, the variance and noise within a class are greatly reduced. Taking advantage of the proxy-based labels, we design an intra-camera contrastive learning~\cite{Chen2020SimCLR} component to jointly tackle multiple camera-specific Re-ID tasks. When compared to the global Re-ID task, each camera-specific task deals with less number of IDs and smaller variance while using more reliable pseudo labels, and therefore is easier to learn. The intra-camera learning enables our Re-ID model to effectively learn discrimination ability within cameras. Besides, we also design an inter-camera contrastive learning component, which exploits both positive and hard negative proxies across cameras to learn global discrimination ability. A proxy-balanced sampling strategy is also adopted to select appropriate samples within each mini-batch, facilitating the model learning further.

In contrast to previous clustering-based methods, the proposed approach distinguishes itself in the following aspects:
\begin{itemize}
	\item Instead of using camera-agnostic clusters, we produce camera-aware proxies which can better capture local structure within IDs. They also enable us to deal with large intra-ID variance caused by different cameras, and generate more reliable pseudo labels for learning. 
	\item With the assistance of the camera-aware proxies, we design both intra- and inter-camera contrastive learning components which effectively learn ID discrimination ability within and across cameras. We also propose a proxy-balanced sampling strategy to facilitate the model learning further. 
	\item Extensive experiments on three large-scale datasets, including Market-1501~\cite{7410490}, DukeMTMC-reID~\cite{zheng2017unlabeled}, and MSMT17~\cite{wei2018person}, show that the proposed approach outperforms both purely unsupervised and UDA-based methods. Especially, on the challenging MSMT17 dataset, we gain $14.3\%$ Rank-1 and $10.2\%$ mAP improvements when compared to the second place. 
\end{itemize}

\section{Related Work}
\subsection{Unsupervised Person Re-ID}
According to whether using external labeled datasets or not, unsupervised Re-ID methods can be grouped into purely unsupervised or UDA-based categories. 

\textbf{Purely unsupervised person Re-ID} does not require any annotations and thus is more challenging. Existing methods mainly resort to pseudo labels for learning. Clustering~\cite{lin2019aBottom,zeng2020hierarchical}, k-NN~\cite{li2018unsupervised,chen2018deepa}, or graph~\cite{ye2017dynamic,wu2019graph} based association techniques have been developed to generate pseudo labels. Most clustering-based methods like BUC~\cite{lin2019aBottom} and HCT~\cite{zeng2020hierarchical} perform in a camera-agnostic way, which can maintain the similarity within IDs but may neglect the intra-ID variance caused by the change of camera views. Conversely, TAUDL~\cite{li2018unsupervised}, DAL~\cite{chen2018deepa}, and UGA~\cite{wu2019graph} divide the Re-ID task into intra- and inter-camera learning stages, by which the discrimination ability learned from intra-camera can facilitate ID association across cameras. These methods generate intra-camera pseudo labels via a sparse sampling strategy, and they need a proper way for inter-camera ID association. In contrast to them, our cross-camera association is straightforward. Moreover, we propose distinct learning strategies in both intra- and inter-camera learning parts.

\textbf{Unsupervised domain adaptation (UDA) based person Re-ID} requires some source datasets that are fully annotated, but leaves the target dataset unlabeled. Most existing methods address this task by either transferring image styles~\cite{Wei2018PTGAN,Deng2018SPGAN,Liu2019ATNet} or reducing distribution discrepancy~\cite{qi2019DA,Wu2019CA} across domains. These methods focus more on transferring knowledge from source to target domain, leaving the unlabeled target datasets underexploited. To sufficiently exploit unlabeled data, clustering~\cite{unsup_clustering, zhai2020ad, ge2020self} or k-NN~\cite{zhong2019invariance} based methods have also been developed, analogous to those introduced in the purely unsupervised task. Differently, these methods either take into account both original and transferred data~\cite{unsup_clustering,zhong2019invariance,ge2020self}, or integrate a clustering procedure together with an adversarial learning step~\cite{zhai2020ad}.

\begin{figure*}[t]
\centering
\centering
\includegraphics[width=0.76\textwidth]{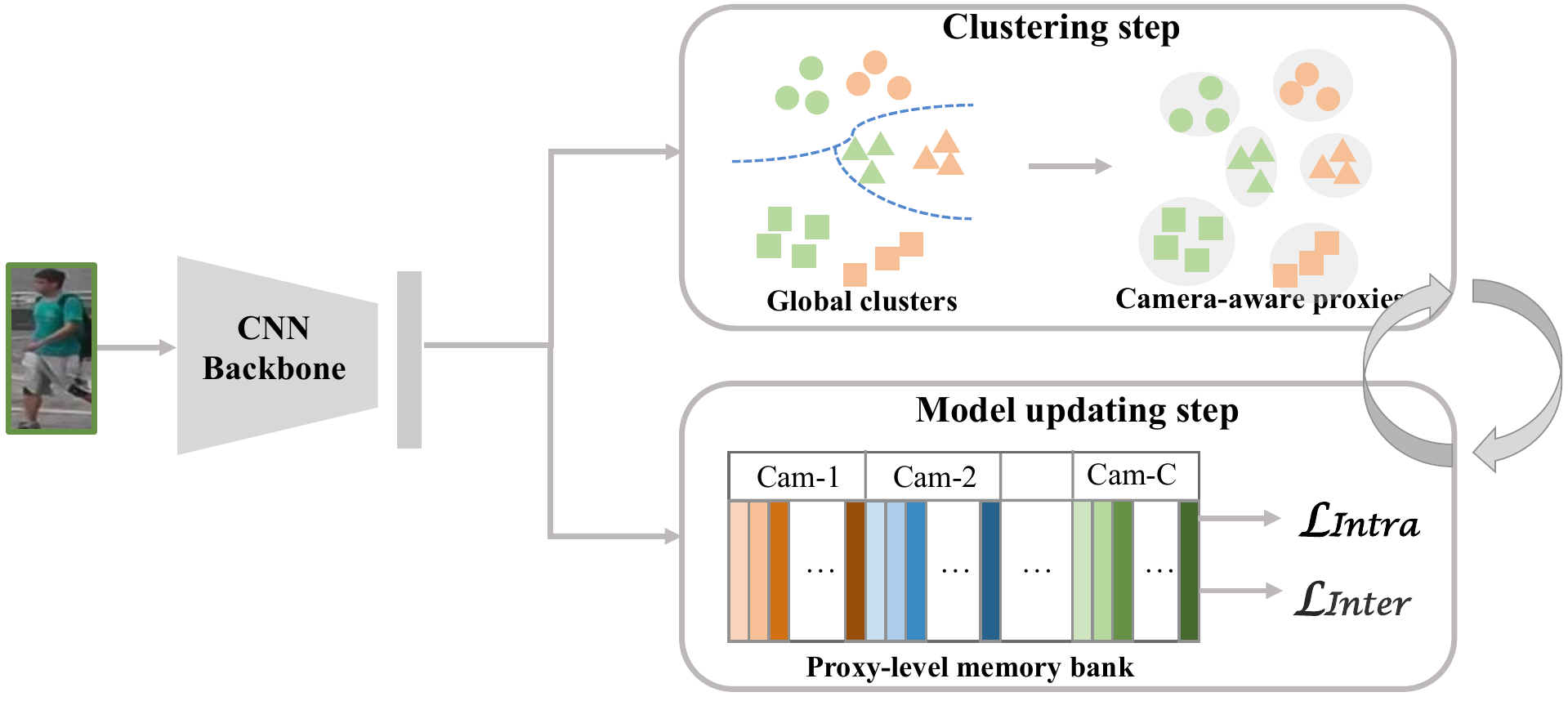} 
\caption{An overview framework of the proposed method. It iteratively alternates between a clustering step and a model updating step. In the clustering step, a global clustering is first performed and then each cluster is split into multiple camera-aware proxies to generate pseudo labels. In the model updating step, intra- and inter-camera losses are designed based on a proxy-level memory bank to perform contrastive learning.}
\label{fig_framework}
\end{figure*}

\subsection{Intra-Camera Supervised Person Re-ID}
Intra-camera supervision (ICS)~\cite{zhu2019intra, qi2019progressive} is a new setting proposed in recent years. It assumes that IDs are independently labeled within each camera view and no inter-camera ID association is annotated. Therefore, how to effectively perform the supervised intra-camera learning and the unsupervised inter-camera learning are two key problems. To address these problems, various methods such as PCSL~\cite{qi2019progressive}, ACAN~\cite{qi2019intra}, MTML~\cite{zhu2019intra}, MATE~\cite{zhu2020intra}, and Precise-ICS~\cite{wang2021} have been developed. Most of these methods pay much attention to the association of IDs across cameras. When taking camera-aware proxies as pseudo labels, our work shares a similar scenario in the intra-camera learning with these ICS methods. Differently, our inter-camera association is straightforward due to the proxy generation scheme. We therefore focus more on the way to generate reliable proxies and conduct effective learning. Besides, the unsupervised Re-ID task tackled in our work is more challenging than the ICS problem.

\subsection{Metric Learning with Proxies}
Metric learning plays an important role in person Re-ID and other fine-grained recognition tasks. An extensively utilized loss for metric learning is the triplet loss~\cite{hermans2017defense}, which considers the distances of an anchor to a positive instance and a negative instance. 
Proxy-NCA~\cite{Attias2017proxy} proposes to use proxies for the measurement of similarity and dissimilarity. A proxy, which represents a set of instances, can capture more contextual information. Meanwhile, the use of proxies instead of data instances greatly reduces the triplet number. Both advantages help metric learning to gain better performance. Further, with the awareness of intra-class variances, Magnet~\cite{Rippel2016multi-center}, MaPML~\cite{Qian2018proxy}, SoftTriple~\cite{qian2019softtriple} and and GEORGE~\cite{sohoni2020no} adopt multiple proxies to represent a single cluster, by which local structures are better represented. 
Our work is inspired by these studies. However, in contrast to set a fixed number of proxies for each class or design a complex adaptive strategy, we split a cluster into a variant number of proxies simply according to the involved camera views, making our proxies more suitable for the Re-ID task.

\section{A Clustering-based Re-ID Baseline}
\label{sec:baseline}
We first set up a baseline model for the unsupervised Re-ID task. As the common practice in the clustering-based methods~\cite{unsup_clustering,lin2019aBottom,zeng2020hierarchical}, our baseline learns a Re-ID model iteratively and, at each iteration, it alternates between a clustering step and a model updating step. In contrast to these existing methods~\cite{unsup_clustering,lin2019aBottom,zeng2020hierarchical}, we adopt a different strategy in the model updating step, making our baseline model more effective. The details are introduced as follows. 

Given an unlabeled dataset $\mathcal{D} = \{x_i\}_{i=1}^{N}$, where $x_i$ is the $i$-th image and $N$ is the image number. We build our Re-ID model upon a deep neural network $f_\theta$ with parameters $\theta$. The parameters are initialized by an ImageNet~\cite{krizhevsky2012imagenet}-pretrained model. When image $x$ is input, the network performs feature extraction and outputs feature $f_\theta(x)$. Then, at each iteration, we adopt DBSCAN~\cite{ester1996density} to cluster the features of all images, and further select reliable clusters by leaving out isolated points. All images within each cluster are assigned with a same pseudo identity label. By this means, we get a labeled dataset $\mathcal{D}' = \{(x_i, \tilde{y}_i)\}_{i=1}^{N'}$, in which $\tilde{y}_i \in \{1, \cdots, Y\}$ is a generated pseudo label. $N'$ is the number of images contained in the selected clusters and $Y$ is the cluster number. 

Once pseudo labels are generated, we adopt a non-parametric classifier~\cite{wu2018memory} for model updating. It is implemented via an external memory bank and a non-parametric Softmax loss. More specifically, we construct a memory bank $\mathcal{K}\in R^{d \times Y}$, where $d$ is the feature dimension. During back-propagation when the model parameters are updated by gradient descent, the memory bank is updated by
\begin{equation}
	\mathcal{K}[j] \leftarrow \mu \mathcal{K}[j] + (1 - \mu) f_\theta(x_i),
	\label{eq:mu}
\end{equation}
where $\mathcal{K}[j]$ is the $j$-th entry of the memory, storing the updated feature centroid of class $j$. Moreover, $x_i$ is an image belonging to class $j$ and $\mu \in [0,1]$ is an updating rate. 

Then, the non-parametric Softmax loss is defined by 
\begin{equation}
\mathcal{L}_{Base} = - \sum_{i=1}^{N'} \log \frac{exp(\mathcal{K}[\tilde{y}_i]^T f_\theta(x_i)/\tau)}{\sum_{j=1}^{Y} exp(\mathcal{K}[j]^T f_\theta(x_i)/\tau)},
\label{eq_base_loss}
\end{equation}
where $\tau$ is a temperature factor. This loss achieves classification via pulling an instance close to the centroid of its class while pushing away from the centroids of all other classes. This non-parametric loss plays a key role in recent contrastive learning techniques~\cite{wu2018memory,zhong2019invariance,Chen2020SimCLR,he2019momentum}, demonstrating a powerful ability in unsupervised feature learning.

\section{The Camera-aware Proxy Assisted Method}
Like previous clustering-based methods~\cite{unsup_clustering,lin2019aBottom,zeng2020hierarchical,zhai2020ad}, the above-mentioned baseline model conducts clustering in a camera-agnostic way. This clustering way may maintain the similarity within each identity class, but neglect the intra-ID variance. Considering that most severe intra-ID variance is caused by the change of camera views, we split each single class into multiple camera-specific proxies. Each proxy represents the instances coming from the same camera. The obtained camera-aware proxies not only capture the variance within classes, but also enable us to divide the model updating step into intra- and inter-camera learning parts. Such a divide-and-conquer strategy facilitates our model updating. The entire framework is illustrated in Figure~\ref{fig_framework}, in which the modified clustering step and the improved model updating step are alternatively iterated.

More specifically, at each iteration, we split the camera-agnostic clustering results into camera-aware proxies, and generate a new set of pseudo labels that are assigned in a per-camera manner. That is, the proxies within each camera view are independently labeled. It also means that two proxies split from the same cluster may be assigned with two different labels. We denote the newly labeled dataset of the $c$-th camera by $\mathcal{D}_c = \{(x_i, \tilde{y}_i, \tilde{z}_i, c_i)\}_{i=1}^{N_c}$. Here, image $x_i$, which previously is annotated with a global pseudo label $\tilde{y}_i$, is additionally annotated with an intra-camera pseudo label $\tilde{z}_i \in \{1, \cdots, Z_c\}$ and a camera label $c_i = c \in \{1, \cdots, C\}$. $N_c$ and $Z_c$ are, respectively, the number of images and proxies in camera $c$, and $C$ is the number of cameras. Then, the entire labeled dataset is $\mathcal{D}''=\bigcup_{c=1}^C \mathcal{D}_c$.

Consequently, we construct a proxy-level memory bank $\mathcal{K}'\in R^{d \times Z}$, where $Z=\sum_{c=1}^C Z_c$ is the total number of proxies in all cameras. Each entry of the memory stores a proxy, which is updated by the same strategy as introduced in Eq. (\ref{eq:mu}) but considers only the images belonging to the proxy. Based on the memory bank, we design an intra-camera contrastive learning loss $\mathcal{L}_{Intra}$ that jointly learns per-camera non-parametric classifiers to gain discrimination ability within cameras. Meanwhile, we also design an inter-camera contrastive learning loss $\mathcal{L}_{Inter}$, which considers both positive and hard negative proxies across cameras to boost the discrimination ability further.

\subsection{The Intra-camera Contrastive Learning}
With the per-camera pseudo labels, we can learn a classifier for each camera and jointly learn all the classifiers. This strategy has the following two advantages. First, the pseudo labels generated from the camera-aware proxies are more reliable than the global pseudo labels. It means that the model learning can suffer less from label noise and gain better intra-camera discrimination ability. Second, the feature extraction network shared in the joint learning is optimized to be discriminative in different cameras concurrently, which implicitly helps the Re-ID model to gain cross-camera discrimination ability. 

Therefore, we learn one non-parametric classifier for each camera and jointly learn classifiers for all cameras. To this end, we define the intra-camera contrastive learning loss as follows.
\begin{equation}
\small
\mathcal{L}_{Intra} = -\sum_{c=1}^{C}\frac{1}{N_c}\sum_{x_i\in \mathcal{D}_c} \log \frac{exp(\mathcal{K}'[j]^T f(x_i)/\tau)}{\sum _{k=A+1}^{A+Z_{c_i}} exp(\mathcal{K}'[k]^T f(x_i)/\tau)}.
\label{eq_intra_loss}
\end{equation}
Here, given image $x_i$, together with its per-camera pseudo label $\tilde{z}_i$ and camera label $c_i$, we set $A = \sum_{c=1}^{c_i-1} Z_c$ to be the total proxy number accumulated from the first to the $c_i-1$-th camera, and $j= A + \tilde{z}_i$ to be the index of the corresponding entry in the memory. $\frac{1}{N_c}$ is to balance the various number of images in different cameras. 

\begin{figure}[t]
\centering
\includegraphics[width=0.45\textwidth]{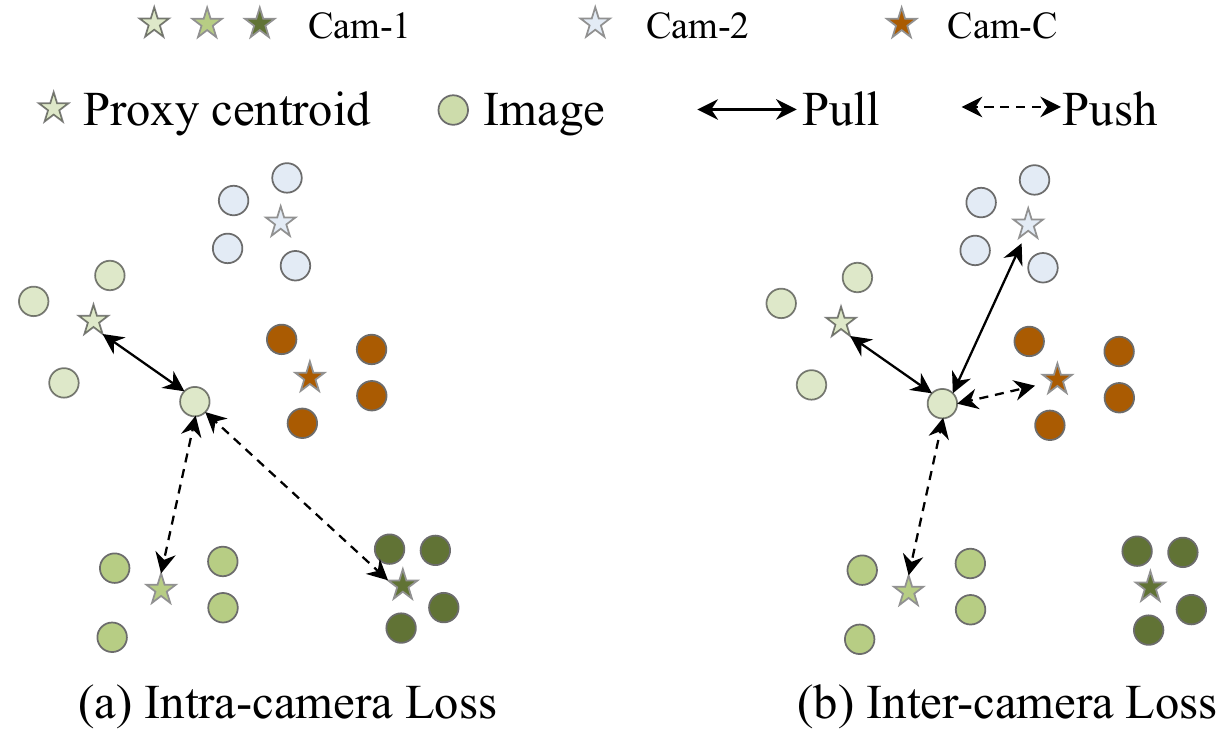}
\caption{Illustration of intra- and inter-camera losses.}
\label{fig_loss}
\end{figure}

This loss performs contrastive learning within cameras. As illustrated in Figure~\ref{fig_loss}(a), this loss pulls an instance close to the proxy to which it belongs and pushes it away from all other proxies in the same camera. 

\subsection{The Inter-camera Contrastive Learning}
Although the intra-camera learning introduced above provides our model with considerable discrimination ability, the model is still weak at cross-camera discrimination. Therefore, we propose an inter-camera contrastive learning loss, which explicitly exploits correlations across cameras to boost the discrimination ability.

Specifically, given image $x_i$, we retrieve all positive proxies from different cameras, which share the same global pseudo label $\tilde{y}_i$. Besides, the K-nearest negative proxies in all cameras are taken as the hard negative proxies, which are crucial to deal with the similarity across identity classes. The inter-camera contrastive learning loss aims to pull an image close to all positive proxies while push away from the mined hard negative proxies, as demonstrated in Figure~\ref{fig_loss}(b). To this end, we define the loss as follows. 
\begin{equation}
\small
\mathcal{L}_{Inter}=-\sum_{i=1}^{N'}\frac{1}{|\mathcal{P}|}\sum_{p \in \mathcal{P}} \log \frac{S(p, x_i)}{\sum \limits_{u \in \mathcal{P}} S(u, x_i) + \sum \limits_{q \in \mathcal{Q}} S(q, x_i)},
\label{eq_global_loss}
\end{equation}
where $\mathcal{P}$ and $\mathcal{Q}$ denote the index sets of the positive and hard negative proxies, respectively. $|\mathcal{P}|$ is the cardinality of $\mathcal{P}$. Moreover, $S(p, x_i) = exp (\mathcal{K}'[p]^T f(x_i) / \tau)$.

\subsection{A Summary of the Algorithm}
The proposed approach iteratively alternates between the camera-aware proxy clustering step and the intra- and inter-camera learning step. The entire loss for model learning is 
\begin{equation}
\mathcal{L} = \mathcal{L}_{Intra} + \lambda \mathcal{L}_{Inter},
\label{eq_overall_loss}
\end{equation}
where $\lambda$ is a parameter to balance two terms. We summarize the whole procedure in Algorithm~\ref{our_algorithm}. 

\begin{algorithm}[h]{
\caption{Camera-aware Proxy Assisted Learning}
\label{our_algorithm}
\hspace*{0.02in} {\bf Input:}
An unlabeled training set $\mathcal{D}$, a DNN model $f_{\theta}$, the iteration number \textit{num\_iters}, the training batches \textit{num\_batches}, momentum $\mu$, and temperature $\tau$; \\
\hspace*{0.02in} {\bf Output:}
Trained model $f_{\theta}$;
\begin{algorithmic}[1]
\For{iter = 1 to \textit{num\_iters}}
    \State Perform a global clustering and remove outliers;
    \State Split clusters into camera-aware proxies, and generate per-camera pseudo labeled dataset $\mathcal{D}''$;
		\State Construct a proxy-level memory bank $\mathcal{K}'$;
    \For{b = 1 to \textit{num\_batches}}
        \State Sample mini-batch images with a proxy-balanced sampling strategy;
        \State Forward to extract the features of the samples;
        \State Compute the loss in Eq.(\ref{eq_overall_loss});
        \State Backward to update model $f_{\theta}$;
        \State Update proxy entries in the memory with the sample features;
    \EndFor
\EndFor
\end{algorithmic}
}
\end{algorithm}

\textbf{A proxy-balanced sampling strategy.} A mini-batch in Algorithm~\ref{our_algorithm} involves an update to the Re-ID model using a small set of samples. It is not trivial to choose appropriate samples in each batch. Traditional random sampling strategy may be overwhelmed by identities having more images than the others. Class-balanced sampling, that randomly chooses $P$ classes and $K$ samples per class as in~\cite{hermans2017defense}, tends to sample an identity more frequently from image-rich cameras, causing ineffective learning for image-deficient cameras. To make samples more effective, we propose a proxy-balanced sampling strategy. In each mini-batch, we choose $P$ proxies and $K$ samples per proxy. This sampling strategy performs balanced optimization of all camera-aware proxies and enhances the learning of \textit{rare} proxies, thus promoting the learning efficacy.

\section{Experiments}
\subsection{Experiment Setting}
\subsubsection{Datasets and metrics.} 
We evaluate the proposed method on three large-scale datasets: Market-1501~\cite{7410490}, DukeMTMC-reID~\cite{zheng2017unlabeled}, and MSMT17~\cite{wei2018person}. 

\textbf{Market-1501}~\cite{7410490} contains 32,668 images of 1,501 identities captured by 6 disjoint cameras. It is split into three sets. The training set has 12,936 images of 751 identities, the query set has 3,368 images of 750 identities, and the gallery set contains 19,732 images of 750 identities. 

\textbf{DukeMTMC-reID}~\cite{zheng2017unlabeled} is a subset of DukeMTMC~\cite{ristani2016performance}. It contains 36,411 images of 1,812 identities captured by 8 cameras. Among them, 702 identities are used for training and the rest identities are for testing. 

\textbf{MSMT17}~\cite{wei2018person} is the largest and most challenging dataset. It has 126,411 images of 4,101 identities captured in 15 camera views, containing both indoor and outdoor scenarios. 32,621 images of 1041 identities are for training, the rest including 82,621 gallery images and 11,659 query images are for testing.

Performance is evaluated by the Cumulative Matching Characteristic (CMC) and mean Average Precision (mAP), as the common practice. For the CMC measurement, we report Rank-1, Rank-5, and Rank-10. Note that no post-processing techniques like re-ranking \cite{Zhong2017reranking} are used in our evaluation.

\subsubsection{Implementation details.} 
We adopt an ImageNet-pretrained ResNet-50~\cite{he2016deep} as the network backbone. Based upon it, we remove the fully-connected classification layer, and add a Batch Normalization (BN) layer after the Global Average Pooling (GAP) layer. The $L_2$ normalized feature is used for the updating of proxies in the memory during training, and also for the distance ranking during inference. The memory updating rate $\mu$ is empirically set to be $0.2$, the temperature factor $\tau$ is $0.07$, the number of hard negative proxies is $50$, and the balancing factor $\lambda$ in Eq. (\ref{eq_overall_loss}) is $0.5$. At the beginning of each epoch (i.e. iteration), we compute Jaccard distance with k-reciprocal nearest neighbors~\cite{Zhong2017reranking} and use DBSCAN~\cite{ester1996density} with a threshold of $0.5$ for the camera-agnostic global clustering. During training, only the intra-camera loss is used in the first 5 epochs. In the remaining epochs, both the intra- and inter-camera losses work together. 

We use ADAM as the optimizer. The initial learning rate is $0.00035$ with a warmup scheme in the first 10 epochs, and is divided by $10$ after each $20$ epochs. The total epoch number is $50$. Each training batch consists of $32$ images randomly sampled from $8$ proxies with $4$ images per proxy. Random flipping, cropping and erasing are applied as data augmentation. 

\begin{figure*}[t]
\centering
\begin{subfigure}{0.29\textwidth}
\centering
\includegraphics[width=1.0\textwidth]{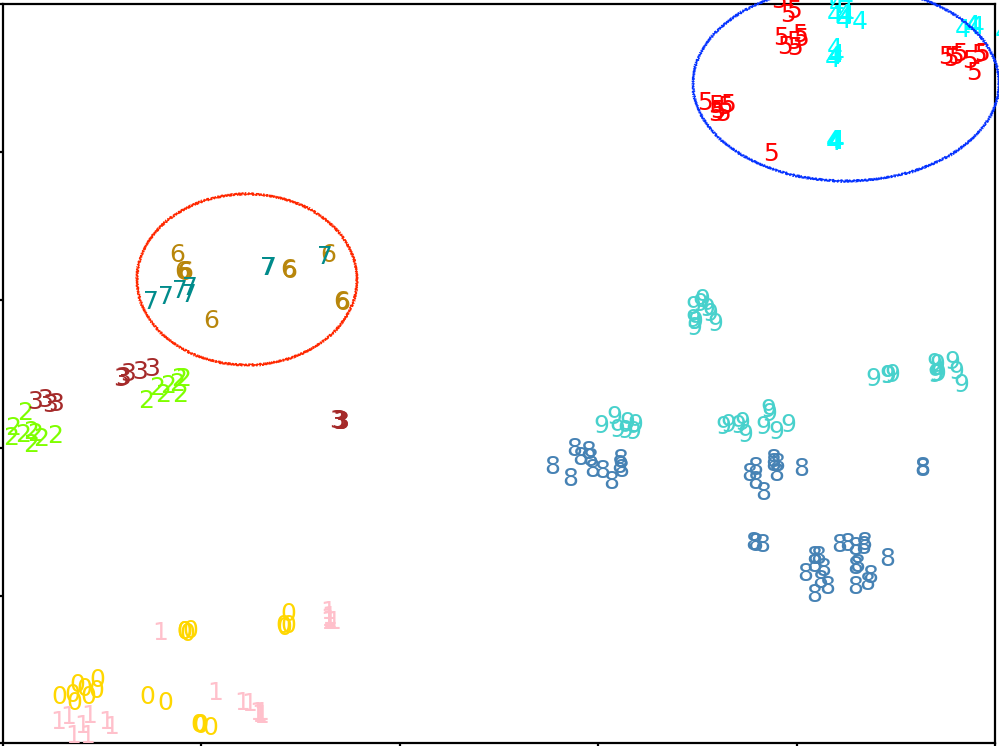} 
\end{subfigure}
\quad
\begin{subfigure}{0.29\textwidth}
\centering
\includegraphics[width=1.0\textwidth]{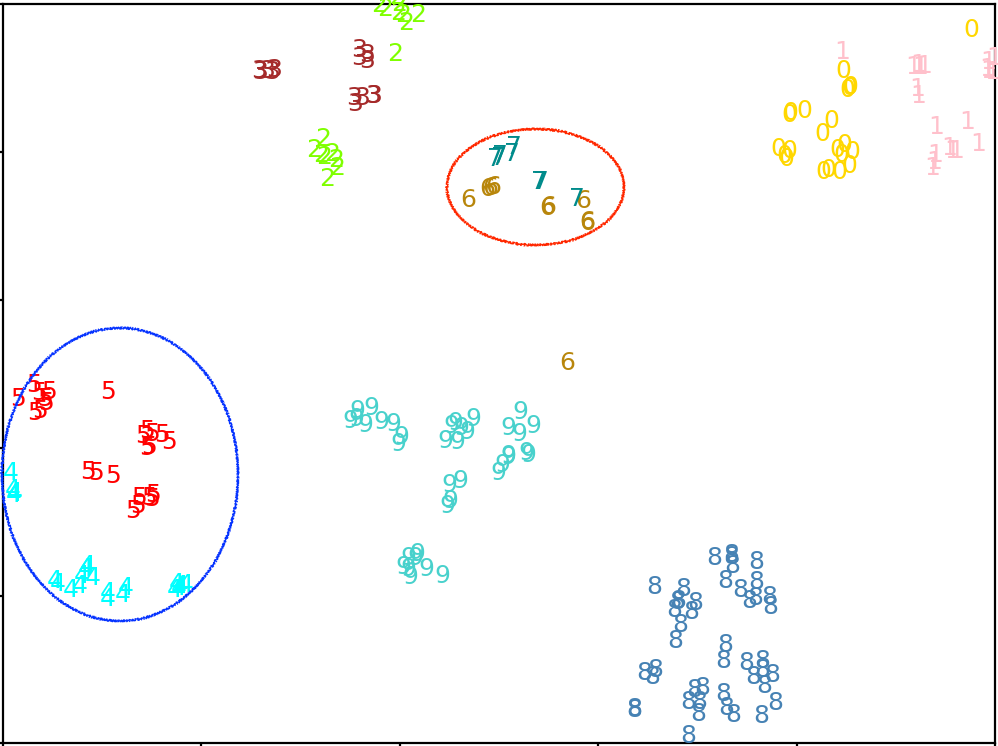}
\end{subfigure}
\quad
\begin{subfigure}{0.29\textwidth}
\centering
\includegraphics[width=1.0\textwidth]{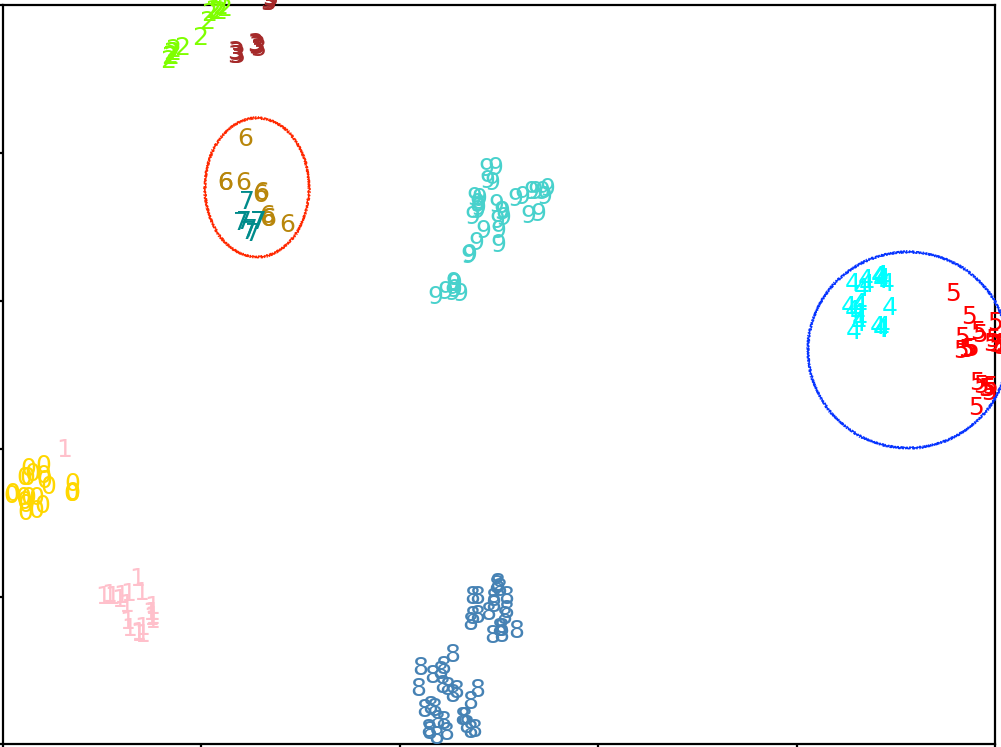}
\end{subfigure}
\\
\begin{subfigure}{0.9\textwidth}
\centering
\includegraphics[width=1.0\textwidth]{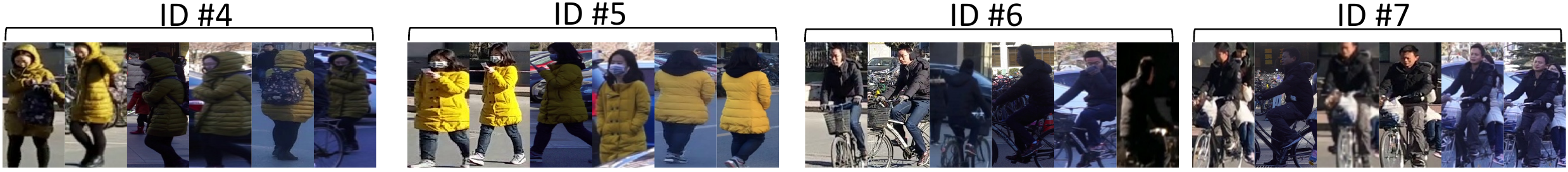}
\end{subfigure}
\caption{T-SNE visualization of features extracted by the models of Baseline, CAP2, and CAP6, respectively shown from left to right in the upper row. Typical examples of IDs \#4-7 are shown at bottom.}
\label{fig_compare_tsne}
\end{figure*}

\subsection{Ablation Studies}
In this subsection, we investigate the effectiveness of the proposed method by examining the intra- and inter-camera learning components, together with the proxy-balanced sampling strategy. For the purpose of reference, we first present the results of the baseline model introduced in section~\ref{sec:baseline}, as shown in Table \ref{ablation_table}. Then, we examine six variants of the proposed camera-aware proxy (CAP) assisted model, which are referred to as CAP1-6.

\begin{table*}[t]
\caption{Comparison of the proposed method and its variants. $\mathcal{L}_{Intra}$ refers to the intra-camera learning, $\mathcal{L}_{Inter}$ is the inter-camera learning, and \textit{PBsampling} is the proxy-balanced sampling strategy. When \textit{PBsampling} is not selected, the model uses the class-balanced sampling strategy.}
\centering
\scalebox{0.89}{
\begin{tabular}{c|ccc|cccc|cccc|cccc}
\hline  
\multirow{2}{*}{Models} & \multicolumn{3}{c|}{Components} &\multicolumn{4}{c| }{Market-1501} & \multicolumn{4}{c|}{DukeMTMC-ReID}  & \multicolumn{4}{c}{MSMT17}  \\
\cline{2-16}  &$\mathcal{L}_{Intra}$ & $\mathcal{L}_{Inter}$ & \textit{PBsampling} & R1 & R5 & R10 & mAP   & R1 & R5 & R10  & mAP   & R1 & R5 & R10 & mAP \\ 
\hline
 Baseline & &  &  &         79.7 & 88.3 & 91.2 & 62.9                  &74.3 & 82.7 & 86.0 & 57.5            & 34.0 & 43.7  & 49.0 & 13.7\\ \hline
CAP1 & \cmark &  &    & 78.7 & 89.3 & 92.9 & 58.9           & 74.0 & 83.7 & 86.6 & 57.0          & 48.6 & 61.7 & 67.1 & 23.0\\ 
CAP2 & \cmark &  & \cmark    & 82.3	& 91.7& 94.1& 64.6           & 76.5 & 86.4 & 89.8 & 60.9          & 51.3 & 64.0 & 69.4 & 24.8\\ 
CAP3 &  & \cmark &      & 89.8 & 95.4 & 97.1 & 75.1        & 76.7 & 84.8 & 86.8 & 59.9          & 66.3 & 76.5 & 80.0 & 34.0 \\
CAP4 &  & \cmark & \cmark     &  91.1 & 96.3 & 97.4 & \textbf{79.9}       & 78.0 & 85.6 & 87.9          & 61.6 & 66.9 & 77.4 & 80.7 & 35.3 \\
CAP5 & \cmark &  \cmark &    & 89.5 & 94.9 & 96.4 & 75.9           & 79.1 & 87.8 & 89.9  &64.5            & 66.7 & 76.9 & 80.5 & 35.1\\ 
CAP6 & \cmark & \cmark & \cmark     & \textbf{91.4} & \textbf{96.3} & \textbf{97.7} & 79.2                 & \textbf{81.1} & \textbf{89.3} & \textbf{91.8} & \textbf{67.3}            & \textbf{67.4} & \textbf{78.0}  & \textbf{81.4} & \textbf{36.9}\\ 

\hline
\end{tabular}
}
\label{ablation_table}
\end{table*}

Compared with the baseline model, the proposed full model (CAP6) significantly boosts the performance on all three datasets. The full model gains $11.7\%$ Rank-1 and $16.3\%$ mAP improvements on Market-1501, and $6.8\%$ Rank-1 and $9.8\%$ mAP improvements on DukeMTMC-ReID. Moreover, it dramatically boosts the performance on MSMT17, achieving $33.4\%$ Rank-1 and $23.2\%$ mAP improvements over the baseline. The MSMT17 dataset is a lot more challenging than the other two datasets, containing complex scenarios and appearance variations. The superior performance on MSMT17 shows that our full model gains an outstanding ability to deal with severe intra-ID variance. In the followings, we take a close look at each component.

\textbf{Effectiveness of the intra-camera learning.}
Compared with the baseline model, the intra-camera learning benefits from two aspects. 1) Each intra-camera Re-ID task is easier than the global counterpart because it deals with less number of IDs and smaller intra-ID variance. 2) Intra-camera learning suffers less from label noise since the per-camera pseudo labels are more reliable. These advantages enable the intra-camera learning to gain promising performance. As shown in Table~\ref{ablation_table}, the CAP1 model which only employs the intra-camera loss, performs comparable to the baseline. When adopting the proxy-based sampling strategy, the CAP2 model outperforms the baseline on all datasets. In addition, we can also observe that the performance drops when removing the intra-camera loss from the full model (CAP4 vs. CAP6), validating the necessity of this component. 

\textbf{Effectiveness of the inter-camera learning.} Complementary to the above-mentioned intra-camera learning, the inter-camera learning improves the Re-ID model by explicitly exploiting the correlations across cameras. It not only can deal with the intra-ID variance via pulling positive proxies together, but also can tackle the inter-ID similarity problem via pushing hard negative proxies away. 
With this component, both CAP5 and CAP6 significantly boost the performance over CAP1 and CAP2 respectively. In addition, we find out that the inter-camera loss alone (CAP3) is able to produce decent performance, and adding the intra-camera loss or sampling strategy boosts performance further.


\textbf{Effectiveness of the proxy-balanced sampling strategy.} The proxy-balanced sampling strategy is proposed to balance the various number of images contained in different proxies. To show that the proxy-balanced sampling strategy is indeed helpful, we compare it with the extensively used class-balanced strategy which ignores camera information. Table~\ref{ablation_table} shows that the models (CAP2, CAP4, and CAP6) using our sampling strategy are superior to the counterparts, validating the effectiveness of this strategy.

\begin{table*}[ht]
\centering
\caption{Comparison with state-of-the-art methods. Both purely unsupervised and UDA-based methods are included. We also provide several fully supervised methods for reference. The first and second best results among all unsupervised methods are, respectively, marked in {\color{red}red} and {\color{blue}blue}. $^\dagger$ indicates an UDA-based method working under the purely unsupervised setting.}
\scalebox{0.89}{
\begin{tabular}{c|c|cccc|cccc|cccc}
\hline  
\multirow{2}{*}{Methods} & \multirow{2}{*}{Reference} &  \multicolumn{4}{c| }{Market-1501} & \multicolumn{4}{c|}{DukeMTMC-ReID}  & \multicolumn{4}{c}{MSMT17}  \\
\cline{3-14}&  & R1 & R5 & R10 & mAP   & R1 & R5 & R10  & mAP   & R1 & R5 & R10 & mAP \\ 
\hline
\multicolumn{13}{l}{\textbf{\textit{Purely Unsupervised}}} \\
BUC~\cite{lin2019aBottom}	& AAAI19        & 66.2 & 79.6 & 84.5  & 38.3              & 47.4 & 62.6 & 68.4 & 27.5         & - & - & - & - \\
UGA~\cite{wu2019graph} & ICCV19 	& 87.2 & - & - & 70.3 & 75.0 & - & - & 53.3 & 49.5 & - & - & 21.7 \\
SSL~\cite{lin2020unsupervised}	& CVPR20      & 71.7 & 83.8 & 87.4 &37.8	        & 52.5 &63.5 &68.9 & 28.6          & - & - & - & - \\
MMCL$^\dagger$~\cite{wang2020unsupervised} & CVPR20       & 80.3 & 89.4 & 92.3 & 45.5      & 65.2 & 75.9 & 80.0  & 40.2      & 35.4 & 44.8 & 49.8 & 11.2\\
HCT~\cite{zeng2020hierarchical}    & CVPR20		& 80.0 & 91.6 & 95.2 & 56.4    & 69.6	& 83.4 & 87.4 & 50.7	         & - & - & - & - \\
CycAs~\cite{wang2020cycas} & ECCV20                  & 84.8 & - & - & 64.8 	                 & 77.9 &- & - & 60.1                    & 50.1 & - & - & {\color{blue}26.7} \\
SpCL$^\dagger$~\cite{ge2020self}          & NeurIPS20       & 88.1 & 95.1 & 97.0 & 73.1      &- & - & - & -            & 42.3 & 55.6 & 61.2 & 19.1 \\
CAP        & This paper                 & {\color{red}91.4} & {\color{red}96.3} & {\color{red}97.7} & {\color{red}79.2}          & {\color{blue}81.1} & {\color{blue}89.3} & {\color{blue}91.8} & {\color{blue}67.3}            & {\color{red}67.4} & {\color{red}78.0}  & {\color{red}81.4} & {\color{red}36.9}\\ 
\hline
\multicolumn{13}{l}{\textbf{\textit{Unsupervised Domain Adaptation}}} \\
PUL~\cite{unsup_clustering}  & TOMM18        & 45.5 &  60.7 &66.7 & 20.5         & 30.0 & 43.4 & 48.5 & 16.4           & - & - & - & - \\
SPGAN~\cite{deng2018similarity}  & CVPR18    & 51.5 & 70.1 & 76.8 & 22.8       & 41.1 & 56.6 & 63.0 & 22.3          & - & - & - & - \\
ECN~\cite{zhong2019invariance}    & CVPR19      & 75.1 & 87.6 & 91.6 & 43.0      & 63.3 & 75.8 & 80.4 & 40.4    & 30.2 & 41.5 & 46.8 & 10.2\\
pMR~\cite{Wang_2020_CVPR}         & CVPR20     & 83.0 & 91.8 & 94.1 & 59.8             & 74.5 & 85.3 & 88.7 & 55.8            & - & - & -  & -\\
MMCL~\cite{wang2020unsupervised} & CVPR20    & 84.4 & 92.8 & 95.0 & 60.4    & 72.4 & 82.9 & 85.0 & 51.4         & 43.6 & 54.3 & 58.9 & 16.2\\
AD-Cluster~\cite{zhai2020ad} & CVPR20 & 86.7 & 94.4 & 96.5 & 68.3 & 72.6 & 82.5 & 85.5 & 54.1 & - & - & - & - \\
MMT~\cite{ge2020mutual}   & ICLR20              &87.7 & 94.9 & 96.9 & 71.2       & 78.0 & 88.8 & 92.5 & 65.1     & 50.1 & 63.9 & 69.8 & 23.3\\
SpCL~\cite{ge2020self}        & NeurIPS20    & {\color{blue}90.3} & {\color{blue}96.2} & {\color{red}97.7} & {\color{blue}76.7}     & {\color{red}82.9} & {\color{red}90.1} & {\color{red}92.5} &{\color{red}68.8}      & {\color{blue}53.1} & {\color{blue}65.8} & {\color{blue}70.5} & 26.5\\
\hline
\multicolumn{13}{l}{\textbf{\textit{Fully Supervised}}} \\
PCB~\cite{sun2018beyond}   & ECCV18   & 93.8 &- &- & 81.6                  &83.3 &- &-   &69.2           &68.2 &- &- &40.4\\ 
ABD-Net~\cite{chen2019abd} & ICCV19 & 95.6 &- &- & 88.3 & 89.0 &- &- & 78.6 & 82.3 & 90.6 &- & 60.8 \\
CAP's Upper Bound       & This paper   & 93.3 & 97.5 & 98.4  & 85.1                 &87.7 & 93.7 & 95.4    &76.0         & 77.1 & 87.4 & 90.8  & 53.7 \\
\hline
\end{tabular}
}
\label{compare_SOTA_table}
\end{table*}   

\textbf{Visualization of learned feature representations.} In order to investigate how each learning component behaves, we utilize t-SNE~\cite{vanDerMaaten2008} to visualize the feature representations learned by the baseline model, the intra-camera learned model CAP2, and the full model CAP6. Figure~\ref{fig_compare_tsne} presents the image features of 10 IDs taken from MSMT17. From the figure we observe that the baseline model fails to distinguish $\#0$ and $\#1$, $\#4$ and $\#5$,  $\#6$ and $\#7$. In contrast, the CAP2 model, which conducts the intra-camera learning only, separates $\#4$ and $\#5$, $\#8$ and $\#9$ better. 
With the additional inter-camera learning component, the full model can distinguish most of the IDs, by greatly improving the intra-ID compactness and inter-ID separability. But it may still fail in some tough cases such as $\#6$ and $\#7$.

\subsection{Comparison with State-of-the-Arts}
In this section, we compare the proposed method (named as CAP) with state-of-the-art methods. The comparison results are summarized in Table~\ref{compare_SOTA_table}.

\textbf{Comparison with purely unsupervised methods.} Five most recent purely unsupervised methods are included for comparison, which are BUC~\cite{lin2019aBottom}, UGA~\cite{wu2019graph}, SSL~\cite{lin2020unsupervised}, HCT~\cite{zeng2020hierarchical}, and CycAs~\cite{wang2020cycas}. Both BUC and HCT are clustering-based, sharing the same technique with ours. Additionally, we also compare with MMCL$^\dagger$~\cite{wang2020unsupervised} and SpCL$^\dagger$~\cite{ge2020self}, two UDA-based methods working under the purely unsupervised setting. From the table, we observe that our proposed method outperforms all state-of-the-art counterparts by a great margin. For instance, compared with the second place method, our approach obtains $3.3\%$ Rank-1 and $6.1\%$ mAP gain on Market, $3.2\%$ Rank-1 and $7.2\%$ mAP gain on Duke, and $17.3\%$ Rank-1 and $10.2\%$ mAP gain on MSMT17.

\textbf{Comparison with UDA-based methods.} 
Recent unsupervised works focus more on UDA techniques that exploit external labeled data to boost the performance. Table~\ref{compare_SOTA_table} presents eight UDA methods. Surprisingly, without using any labeled information, our approach outperforms seven of them on both Market and Duke, and is on par with SpCL. On the challenging MSMT17 dataset, our approach surpasses all methods by a great margin, achieving $14.3\%$ Rank-1 and $10.4\%$ mAP gain when compared to SpCL.  


\textbf{Comparison with fully supervised methods.} Finally, we provide two fully supervised method for reference, including one well-known method PCB~\cite{sun2018beyond} and one state-of-the-art method ABD-Net~\cite{chen2019abd}. We also report the performance of our network backbone trained with ground-truth labels, which indicates the upper bound of our approach. We observe that our unsupervised model (CAP) greatly mitigates the gap with PCB on all three datasets. Besides, there is still room for improvement if we could improve our backbone via integrating recent attention-based techniques like ABD-Net.

\section{Conclusion}
In this paper, we have presented a novel camera-aware proxy assisted learning method for the purely unsupervised person Re-ID task. Our method is able to deal with the large intra-ID variance resulted from the change of camera views, which is crucial for a Re-ID model to improve performance. With the assistance of camera-aware proxies, our proposed intra- and inter-camera learning components effectively improve ID-discrimination within and across cameras, as validated by the experiments on three large-scale datasets. Comparisons with both purely unsupervised and UDA-based methods demonstrate the superiority of our method.

\bibliography{reference_v3}

\end{document}